# Automated segmentation of pediatric neuroblastoma on multi-modal MRI: Results of the SPPIN challenge at MICCAI 2023

Myrthe A.D. Buser[1]
Dominique. C. Simons[1]
Matthijs. Fitski, PhD[1]
Marc H.W.A. Wijnen[1]
Annemieke S. Littooij[1,2]
Annemiek H. ter Brugge[1]
Iris N. Vos[3]
Markus H.A. Janse[3]
Mathijs de Boer[3]
Rens ter Maat[3]
Junya Sato[4]
Shoji Kido[4]
Satoshi Kondo[5]
Satoshi Kasai[6]
Marek Wodzinski[7,8]
Henning Muller[8,9]
Jin Ye[10]
Junjun He[10]
Yannick Kirchhoff[11]
Maximilian R. Rokkus[11]
Gao Haokai[12]
Su Zitong[12]
Matías Fernández-Patón[13]
Diana Veiga-Canuto[13]
David G. Ellis[14]
Michele R. Aizenberg[14]
Bas H.M. van der Velden[3,15]
Hugo Kuijf[3]
Alberto de Luca[3]
Alida F. W. van der Steeg[1]

[1] Princess Máxima Center for Pediatric Oncology, Utrecht, The Netherlands
[2] Department of Radiology, University Medical Center Utrecht, Utrecht, the Netherlands.
[3] Image Sciences Institute, University Medical Center Utrecht, Utrecht University, Utrecht, The Netherlands
[4] Department of Artificial Intelligence in Diagnostic Radiology, Osaka University Graduate School of Medicine, Japan
[5] Muroran Institute of Technology, Hokkaido, Japan
[6] Niigata University of Health and Welfare, Niigata, Japan
[7] AGH University of Krakow, Krakow, Poland
[8] HES-SO Valais-Wallis, Valais, Switzerland
[9] University of Geneva, Geneva, Switzerland
[10] Shanghai Artificial Intelligence Laboratory, Shanghai, China
[11] Cancer Research Center, Heidelberg, Germany
[12] South China Normal University, Guangzhou, China
[13] La Fe Health Research Institute, Valencia, Spain
[14] University of Nebraska Medical Center, Nebraska, United States of America
[15] Wageningen Food Safety Research, Wageningen, The Netherlands

**Abstract**

*Introduction:* Surgery plays an important role within the treatment for neuroblastoma, a common pediatric cancer. This requires careful planning, often via magnetic resonance imaging (MRI)-based anatomical 3D models. However, creating these models is often time-consuming and user dependent. We organized the Surgical Planning in PedIatric Neuroblastoma (SPPIN) challenge, to stimulate developments on this topic, and set a benchmark for fully automatic segmentation of neuroblastoma on multi-model MRI.

*Methods:* The challenge started with a training phase, where teams received 78 sets of MRI scans from 34 patients, consisting of both diagnostic and post-chemotherapy MRI scans. Then, in the preliminary test phase, the team algorithms could be tested on 7 scans belonging to 3 patients. The final test phase, consisting of 18 MRI sets from 9 patients, determined the ranking of the teams. Ranking was based on the Dice similarity coefficient (Dice score), the $95^{th}$ percentile of the Hausdorff distance (HD95) and the volumetric similarity (VS). The SPPIN challenge was hosted at MICCAI 2023 as a one-time event.

*Results:* The final leaderboard consisted of 9 teams. The highest-ranking team achieved a median Dice score 0.82, a median HD95 of 7.69 mm and a VS of 0.91, utilizing a large, pre-trained network called STU-Net. A significant difference for the segmentation results between diagnostic and post-chemotherapy MRI scans was observed (Dice = 0.89 vs Dice = 0.59, P = 0.01) for the highest-ranking team.

*Conclusion:* SPPIN is the first medical segmentation challenge in extracranial pediatric oncology. The highest-ranking team used a large pre-trained network, suggesting that pre-training can be of use in small, heterogenous datasets. Although the results of the highest-ranking team were high for most patients, segmentation especially in small, pre-treated tumors was insufficient. Therefore, more reliable segmentation methods are needed to create clinically applicable models to aid surgical planning in pediatric neuroblastoma.

# 1. Introduction

Neuroblastoma is one of the most common extra-cranial solid tumors in children [1]. The incidence of neuroblastoma is about 11 per million children under the age of 15 years old [2]. Neuroblastoma accounts for 15% of cancer related deaths in children [1] [3]. Treatment includes chemotherapy, immunotherapy, surgery, and radiotherapy. While the specific treatment strategy depends on multiple factors such as age, tumor biology, and image defined risk factors chemotherapy followed by surgery is one of the mainstays of treatment [1], [4]. Surgery is typically performed to debulk the tumor, aiming to achieve local control by removing at least 95% of tumor tissue [5]. The majority of neuroblastomas are located in the abdomen where surgical debulking can be challenging, due to adherence to important abdominal structures, such as spleen, liver, kidneys and ureter, and abdominal vessels such as aorta, vena cava and renal vessels in the vicinity of the tumor [5], [6].

To safely remove as much tumor tissue as possible without damaging important abdominal structures, it is essential for the surgeon to understand the position of the neuroblastoma in relation to the important abdominal structures. Currently, pre-operative imaging such as magnetic resonance imaging (MRI) is used to investigate the anatomical situation of the patient. In other pediatric tumors, creating 3D models of the tumor in relation with other important structures from the MRI images showed to increase the anatomical understanding of the surgeon during surgical planning, potentially leading to faster procedures with better surgical outcomes [7], [8], [9]. However, creating these 3D models is a user-dependent and time-consuming process, as manual or semi-automatic delineation of both the neuroblastoma and the important abdominal structures is required. [10]. Automating the segmentation (i.e., delineation) of neuroblastoma can increase accessibility, applicability, and reliability of pre-operative 3D models. However, automated segmentation of neuroblastoma is not trivial and not yet common practice. Firstly, neuroblastoma is a rare cancer only affecting the pediatric population, thus limiting the size of available datasets. Secondly, the location, size, shape, and image characteristics of the tumor can vary greatly between patients due to the heterogeneous biology of neuroblastoma [11]. Lastly, chemotherapy-induced changes often decrease the size of the tumor and visibility of the tumor boundaries, making especially the pre-operative segmentation of neuroblastoma from MRI challenging.

Automating the creation of the pre-operative 3D models increases usability while making this technique accessible for more patients. Deep learning can be used to automate segmentation of pediatric oncology on MRI [12], [13], [14], [15]. However, only two studies focus on segmentation of neuroblastoma using deep learning [10], [16]. The method described in these two papers was able to segment tumors with a high accuracy. However, the focus was on diagnostic MRI scans. Therefore, it is unclear if this is generalizable to application in pre-operative planning of pre-treated tumors.

Due to the high complexity of this problem combined with limited research, we wanted to provide a benchmark for fully automatic neuroblastoma segmentation, in both diagnostic as post-chemo MRI scans. To utilize a wide range of knowledge within the medical imaging community, we decided to organize the Surgical Planning in PedIatric Neuroblastoma (SPPIN) challenge. The challenge was hosted online in 2023, with a concluding, in person challenge session in conjunction with the 26th International Conference on Medical Image Computing and Computer Assisted Intervention (MICCAI) in Vancouver, Canada.

Our challenge focused on fully automated neuroblastoma segmentation using multi-modal MRI scans during multiple stages in the treatment process, with a focus on accurate segmentation of post-chemotherapy scans [17]. The aim of this paper is to describe the challenge set-up, explain methods of the participating teams, and provide an objective assessment of the performance of the methods with an eye on their potential future translation to support presurgical planning. The structure of the paper follows the guidelines for transparent reporting of the Biomedical Image Analysis ChallengeS (BIAS) initiative [18].

## 2. Methods
### 2.1 Challenge design

The SPPIN challenge was organized to investigate automatic segmentation of neuroblastoma on MRI, with the aim of using this for surgical planning. The challenge utilized multi-modal, unregistered MRI data to simulate the clinical situation. Although the aim of this challenge was the segmentation of the pre-operative scan, scanning moments before surgical removal of the neuroblastoma were included, to increase the dataset size. This included the diagnostic scans of chemotherapy-naïve patients. All MRI sequences scanned at the same moment (section 2.2.1) will be referred to as one scan, so multiple scans can belong to the same patient, of which an example can be seen in Figure 1. For each scan the patient number and scanning date were provided, but no additional (clinical) information was provided to the teams.

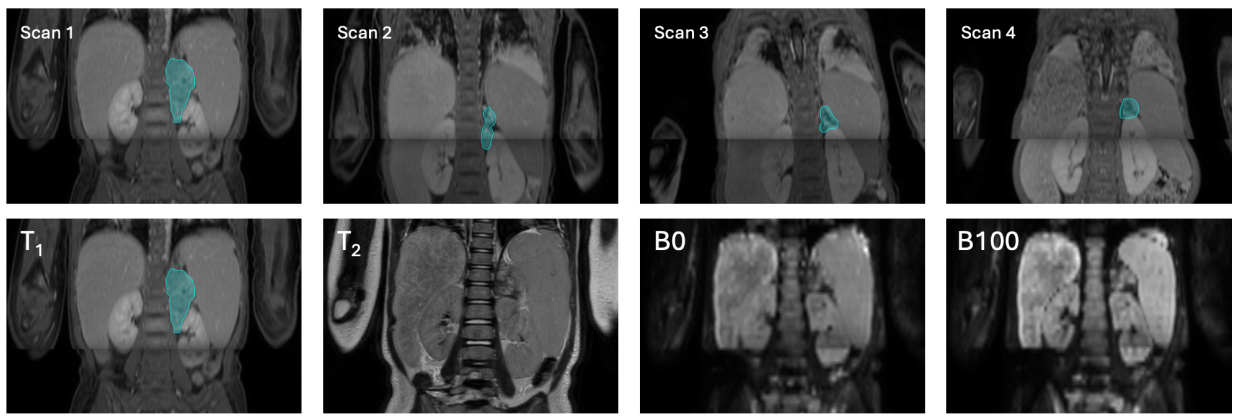

*Figure 1: Example of multiple scans belonging to one patient. The top row depicts the several scanning moments throughout the treatment process: with scan 1 being the diagnostic scan, scan 2 and 3 scans during chemotherapy at different time moments, and scan 4 being the pre-operative scan at the end of the chemotherapy. The bottom row shows the four MRI scans belonging to diagnostic scan. From left to right: the $T_1$-weighted contrast enhanced scan, the $T_2$-weighted scan, the DWI scan (b = 0s/mm$^2$) and the DWI scan (b = 100 s/mm$^2$). The ground truth is depicted on the $T_1$-weighted contrast enhanced scan, where the ground truth was created on, in blue.*

The SPPIN challenge was organized by the Princess Máxima Center for Pediatric Oncology and ran as a satellite event at MICCAI 2023 on October 8[th,] 2023. The peer-reviewed challenge design was previously published [17]. Teams with members of the organizers' institute were not allowed to participate. Only the teams that presented their work at the SPPIN challenge session and provided an overview of their method will be discussed in this article. Detailed information about the challenge is available at https://sppin.grand-challenge.org/.

The SPPIN challenge was hosted on Grand Challenge. Participating teams submitted their automated segmentation methods as a docker container, for which instructions were available [19]. After uploading their method, the segmentation and evaluation were performed automatically within the Grand Challenge platform. Per restrictions of Grand Challenge, the computation time for one scan was limited to 20 minutes. Only fully automated methods were allowed. Use of external data to (pre-)train or fine-tune the used methods was only allowed when a publicly available dataset was used and cited.

The challenge consisted of three phases. The training phase ran from April 14th, 2023, to September 1st, 2023. Data for the training phase was available after signing a data release form. The preliminary test phase ran from May 1st, 2023, to September 1st, 2023. In the preliminary test phase teams could test their method, to ensure that the method behaved as expected. Teams had 5 attempts to submit their method, and the best scoring method compared to the previous attempts was directly posted to the live leaderboard. The final test phase ran from the August 14th, 2023, to September 1st, 2023. In the final test phase, teams only had one attempt, and the results of this leaderboard were hidden until the SPPIN challenge session on October 8th, 2023. This structure of two different test phases ensured that teams could test and debug their methods without being able to optimize their results specifically for the final leaderboard. The SPPIN dataset will remain available upon request with the corresponding author.

## 2.2 Challenge data sets

We retrospectively included neuroblastoma patients aged 0-18 treated in the Princess Máxima Center for Pediatric Oncology during the period of July 2018 to October 2022. For all patients, informed consent was present (n = 93). Patients were excluded when the imaging sequences were not complete (n = 18), only post-operative scans were present (n = 10) or if the manual delineation did not pass the quality check due to time constraints while organizing the challenge (n = 19). An overview of patient inclusion can be seen in Figure 2. Clinical characteristics were collected for analysis. All patients received treatment according to the Dutch Childhood Oncology Group (DCOG) NBL 2009 protocol [20].

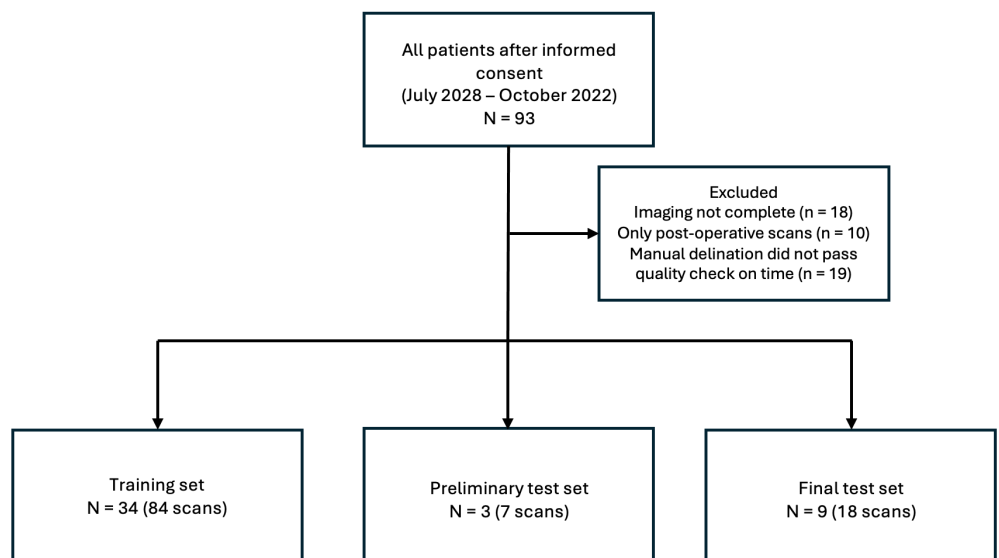

*Figure 2: Overview of patient inclusion. The training set refers to the set of scans the teams received at the beginning at the challenge. The preliminary test set was used as check for the teams, with results being posted directly on the live leaderboard. The final test set was used to determine the final leaderboard and winners of the SPPIN challenge.*

### 2.2.1 Magnetic Resonance Imaging

MRI scanning was performed on a 1.5T unit (Ingenia; Philips Medical Systems, Best, The Netherlands) [20]. The imaging included fat-suppressed $T_1$-weighted images with and without intravenous contrast (Gadovist, Bayer Pharma, Berlin, Germany, 0.1 mmol/kg body weight), a 3D $T_2$-weighted image, and diffusion weighted images (DWI) (Table 1). Scans were pseudomized and exported to NIFTI. No further preprocessing or registration was conducted.

Table 1: Magnetic resonance imaging scanning parameters. The Z dimension of the scans is variable.

| Sequence | $T_1$-weighted gradient echo | $T_2$-weighted spin echo | Diffusion weighted imaging |
|---|---|---|---|
| **Repetition time (ms)** | 6.1 | 458 | 2537 |
| **Echo time (ms)** | 2.9 | 90 | 76.4 |
| **Voxel size (mm³)** | 0.71 x 0.71 x 3 mm | 0.833 x 0.833 x 1.15 | 1.39 x 1.39 x 5 |
| **Dimensions (voxels)** | 560 x 560 x Z | 480 x 480 x Z | 288 x 288 x Z |
| **b-values (s/mm²)** | N/A | N/A | 0, 100 |
| **Contrast** | Gadovist, 0.1 mmol/kg body weight | - | - |

### 2.2.2 Manual annotation

Scans were manually annotated by delineating the neuroblastoma using a custom tool created in MeVisLab [21]. Five trained technical medicine students performed the delineation under direct supervision of two experienced technical physicians. The delineation protocol was cocreated and approved by a dedicated pediatric radiologist (Appendix A).

### 2.2.3 Dataset splits

The included patients were split into train (n = 34 patients with 78 scans), preliminary test (n = 3 patients with 7 scans), and final test set (n = 9 patients with 18 scans) on a patient level. After closing of the challenge, 2 patients were removed from the final test set due to clinical inconsistencies, the reported numbers are consistent with the final analysis. Baseline characteristics are displayed in Table 2.

Table 2: Baseline characteristics of the three challenge datasets. The datasets were split on a patient level.

|  | Training set (n = 34) | Preliminary test set (n = 3) | Final test set (n = 9) |
|---|---|---|---|
| **Number of included scans** | 84 | 7 | 18 |
| **Age in years at diagnosis (median, min-max)** | 2.5 (0-11) | 2 (0 – 3) | 2 (0 – 12) |
| **Gender** | Male = 17 | Male = 2 | Male = 6 |
|  | Female = 17 | Female = 1 | Female = 3 |
| **Tumor volume in mL (median, min-max)** | 53.48 (2.03 – 1249.7) | 7.7 (4.28 – 304.7) | 51.1 (4.9 – 745.4) |

### 2.2.4 Observer variability

To investigate the observer variability, three separate technical medicine students segmented the final test set independently from the original observers. Before starting, they got a section of the original training set to practice (n = 3 patients with 8 scans). These students used the same segmentation protocol and were also under direct supervision of two experienced researchers. The inter-observer variability was determined by calculating the Dice similarity coefficient (Dice score). The original segmentations, as created by one of the five students, were considered as created by one observer. The segmentations created by the three additional students were separately compared against the original segmentations in the test set to obtain the inter-observer variability.

### 2.3 Assessment method
### 2.3.1 Automated evaluation
For each submission, the Dice similarity coefficient (Dice score), the $95^{th}$ percentile of the Hausdorff distance (HD95) and the volumetric similarity (VS) were calculated and ranked [22]. These rankings were combined to get to the final ranking, with the ranking based on the Dice score as tie breaker in case of a tied ranking. Due to the nature of the challenge design, no missing results were present. Empty segmentations were given a HD95 of infinite (in the statistics depicted as NaN). Evaluation and statistical analysis were performed in Python 3.8. The Python code used for the automated evaluation of the segmentations can be referenced at the challenge Github page [23].

### 2.3.2 Statistical analysis
To determine the effect of the two important clinical factors, tumor size and the effect of treatment, statistical analysis of these parameters were performed. Kruskal-Wallis tests were used to determine the difference in segmentation outcomes between diagnostic and post-chemotherapy scans for each team. Furthermore, scatterplots were created for each team to investigate the effect of tumor size on the Dice score.

## 3. Results
### 3.1 Inter-observer scores
The inter-observer Dice scores between the original observer (one of the five initial students) and the three additional observers can be seen in Figure 3. The median inter-observer Dice score across all three observers was 0.77 (min - max: 0.47 – 0.92). The complete overview of the inter-observer Dice scores can be seen in Appendix B. Observer 2 had the highest spread of observer variation, with one totally incorrect tumor segmentation resulting in a Dice score of 0.

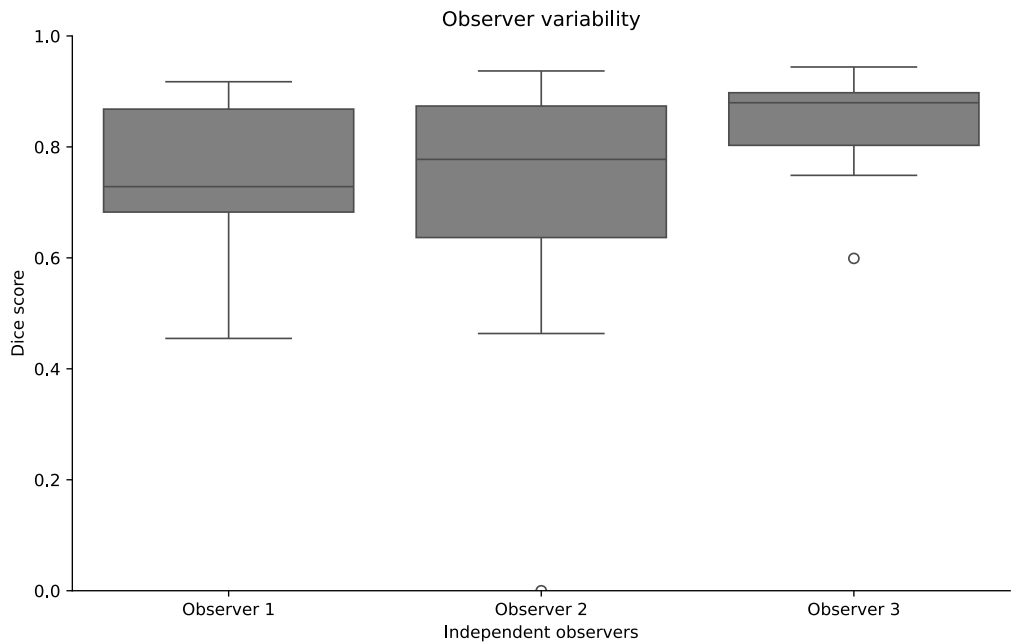

*Figure 3: The distribution of the inter-observer Dice scores. The original segmentations were used as baseline to calculate the inter-observer scores with. These segmentations were created by one of five students. Three additional students, as reflected in this figure, performed the segmentation to obtain the inter-observer scores.*

### 3.2 Challenge submission and participating teams
For the preliminary test phase, 24 valid methods were submitted by 11 teams. For the final test phase, 12 teams submitted a segmentation method resulting in 12 valid methods. This paper includes the 9 teams that submitted both a valid method in the final test phase, and provided a description of the used method. The key characteristics of the team methods are reflected in Table 3 and section 3.2.1.

Table 3: overview of the used segmentation method of each team, from the highest to the lowest ranking team. If an entry was reported but not applied, this was denoted by ' – ' . If an entry was not reported, it was denoted as Not Reported (N.R.). $T_1$ = $T_1$-weighted contrast enhanced scan, $T_2$ = $T_2$-weighted scan, DWI_b0 = diffusion weighted image, b-value 0, DWI_b100 = diffusion weighted image, b-value 100.

| Team | Network | Input | Pre-training | Pre-processing | Patch size | Data augmentation | Loss | Post-processing |
|---|---|---|---|---|---|---|---|---|
| **Blackbean** | Scalable and Transferable U-Net | $T_1$ | TotalSegmentor dataset | nnU-Net default | Yes, size not specified | Yes | Dice loss | N.R. |
| **jishenyu** | nnU-Net | $T_1$, $T_2$, DWI_b0, DWI_b100 | Pre-trained nnU-Net weights | Registration of input to $T_1$ | N.R. | Yes | Dice + cross entropy loss | N.R. |
| **Ouradiology** | nnU-Net | $T_1$ | N.R. | nnU-Net default | 64x288x288 | N.R. | nnU-Net default | nnU-Net default |
| **Drehimpuls** | nnU-Net | $T_1$, $T_2$, DWI_b0, DWI_b100 | N.R. | Resampling input to $T_1$ space, z-score normalization | 128x128x128 | N.R. | Fbeta + cross entropy loss | Fallback network |
| **SK** | 2.5D U-Net, with as encoder EfficientNet | $T_1$, $T_2$ | N.R. | Resampling input to $T_1$ space | N.R. | - | Dice + cross entropy loss | N.R. |
| **AGHSSO** | ResUNet | $T_1$, $T_2$, DWI_b0, DWI_b100 | N.R. | Resampling input to $[224^3]$, normalization | Whole image | Yes | Soft + focal loss | Delete connected components < 20 voxels |
| **UNMC** | DynUNet | $T_1$, $T_2$, DWI_b0, DWI_b100 | N.R. | Registration of input $T_1$, foreground cropping, z-score intensity normalization, linear resampling [192x192x192] | N.R. | Yes | Dice loss | Selection of largest component |
| **SPPIN_SCNU** | UNETR | $T_1$ | N.R. | N.R. | N.R. | Yes | N.R. | N.R. |
| **GIBI230** | nnU-Net | $T_2$ | N.R. | Resampling to [0.695x0.695x8] mm voxel size, z-score normalization | N.R. | N.R. | Dice loss | N.R. |

### 3.2.1 Methods of participating teams

Below, all methods of the participating teams are described in order of their ranking (from highest to lowest). An overview of the used methods can be seen in Table 3.

Blackbean: A Scalable and Transferable U-Net (STU-Net), which is based on nnU-Net, was pretrained on the large dataset TotalSegmentor (1204 CT images with 104 annotated structures) to create this teams segmentation method [24], [25], [26]. As input, this team used $T_1$-weighted contrast enhanced images only. Data augmentation during training included additive brightness, gamma, rotation, scaling, mirror and elastic deformation.

Jishenyu: The segmentation pipeline of this team started with registration of all sequences to the $T_1$-weighted scan. Next, they used a $T_1$-weighted contrast enhanced fill strategy for enhancing tumor visibility by replacing lower intensity pixels of the $T_2$-weighted and diffusion scan with higher intensity values from the corresponding $T_1$-weighted contrast enhanced scan. Data augmentation was used to enhance the dataset variability: Gaussian blur and noise, brightness multiplicate, contrast augmentation, simulation of low resolution, gamma transform, and mirror transform. The segmentation was performed using a pre-trained nnU-Net, for which details were not specified [26].

Ouradiology: This team proposed segmentation based on a nnU-Net, with as input only the $T_1$-weighted images [26]. After training with 5-fold cross validation, their final model was an ensemble of models as determined by nnU-Net.

Drehimpuls: After testing several different nnU-Net algorithms (including the addition of residual connections in the encoder), they showed no enhanced performance compared to a standard nnU-Net [26]. As loss, the $F_\beta$ was used instead of the standard Dice loss [27]. A fallback network was fine-tuned on all scans that had a Dice score of < 0.5 during 5-fold cross validation. This fallback network was used as a postprocessing step in case the first network predicted < 1000 foreground pixels.

SK: A 2.5D convolutional neural net (CNN), using consecutive slices in a transverse plane, with EfficientNet as encoder was used by this team [28]. After optimalisation, they found that using 5 slices was the best performing depth for the 2.5D CNN. They used $T_1$- contrast enhanced and $T_2$-weighted images, which were resampled but not registered.

AGHSSO: Heavy data augmentation and the use of all four input sequences was central to this team's segmentation method. Using affine transformations, random intensity transformation, Gaussian noise, flipping along all axes, motion artifacts, anisotropy transformation, Gaussian blurring, and offline elastic transformations, they created a total of 10000 training cases. They used a 3D Res-Net as architecture, trained with a combination of Soft Dice Loss and Focal Loss. In postprocessing, all connected components of 20 voxels or less were removed.

UNMC: This team used all four input sequences for their method, registered by a "QuickRigid" registration to the $T_1$-weighted contrast enhanced image [29]. The DynUNet model from MONAI was used as segmentation method [30]. Before training, they performed a background cropping (voxels with an intensity below the 90th percentile). Image normalization and resampling was also performed. During training random flips, rotations, scaling, and shifting of intensity values were used as augmentation. As postprocessing, the largest connected component of the prediction was selected.

SPPIN_SCNU: The method of this team only used $T_1$-weighted contrast enhanced images as input. Their method was based on a U-Net with a transformer as encoder, called UNETR [31].

GIBI230: A previously developed neuroblastoma segmentation method, developed as part of the PRIMAGE project, was used by this team [10], [16]. This is the only team that used the $T_2$-images as input. This team used their previously developed method on our dataset, without any finetune training specifically for our dataset.

### 3.3 Metric values and rankings

In Table 4, the metrics for each team are presented. The highest scoring team had a median Dice score of 0.82, a median HD95 of 7.69mm and median VS of 0.91. The lowest ranked team achieved a median Dice score of 0.21, a median HD95 of 63.41mm and a VS of 0.31. The distribution of the scores can be seen in Figure 4. In Figure 5, an overview of the tumors for which the best and worst Dice scores were observed is provided. Both the tumor with the highest Dice score from the winning team as the tumor with the highest total Dice score were depicted, as these were not the same tumors. The same holds for the lowest Dice scores.

Table 4: The scores and rankings of the participating teams. Dice = Dice similarity coefficient. HD95 = the 95th percentile of the Hausdorff distance. VS = volumetric similarity.

| Team name | Median Dice [min-max] | Ranking Dice | Median HD95 (mm) | Ranking HD95 | Median VS | Ranking VS | Final ranking |
|---|---|---|---|---|---|---|---|
| **Blackbean** | 0.82 [0.00 – 0.93] | 1 | 7.69 [2.82 – 127.35] | 1 | 0.91 [0.16 – 0.99] | 1 | **1** |
| **Jishenyu** | 0.79 [0.00 – 0.93] | 3 | 13.19 [2.83 – 146.91] | 2 | 0.86 [0.01 – 1.00] | 2 | **2** |
| **Ouradiology** | 0.80 [0.00– 0.94] | 2 | 15.91[2.83 – 127.82] | 3 | 0.85 [0.04 – 0.99] | 4 | **3** |
| **Drehimpuls** | 0.77 [0.00 – 0.91] | 4 | 20.71[3.16 – 129.65] | 4 | 0.85 [0.02 – 1.00] | 3 | **4** |
| **SK** | 0.57 [0.00 – 0.83] | 5 | 32.32 [5.48 – 128.51] | 6 | 0.77 [0.09 – 0.98] | 5 | **5** |
| **AGHSSO** | 0.48 [0.00 – 0.87] | 6 | 24.11[6.08 – 132.51] | 5 | 0.64 [0.03 – 0.99] | 7 | **6** |
| **UNMC** | 0.40 [0.00 – 0.76] | 7 | 36.26 [11.58 – 116.97] | 7 | 0.69 [0.07 – 0.91] | 6 | **7** |
| **SPPIN_SCNU** | 0.24 [0.00 – 0.61] | 8 | 93.54 [27.20 – 274.0] | 9 | 0.58 [0.03 – 0.99] | 8 | **8** |
| **GIBI230** | 0.21 [0.00 – 0.89] | 9 | 63.41 [5.48 – 170.38] | 8 | 0.31 [0.00 – 0.96] | 9 | **9** |

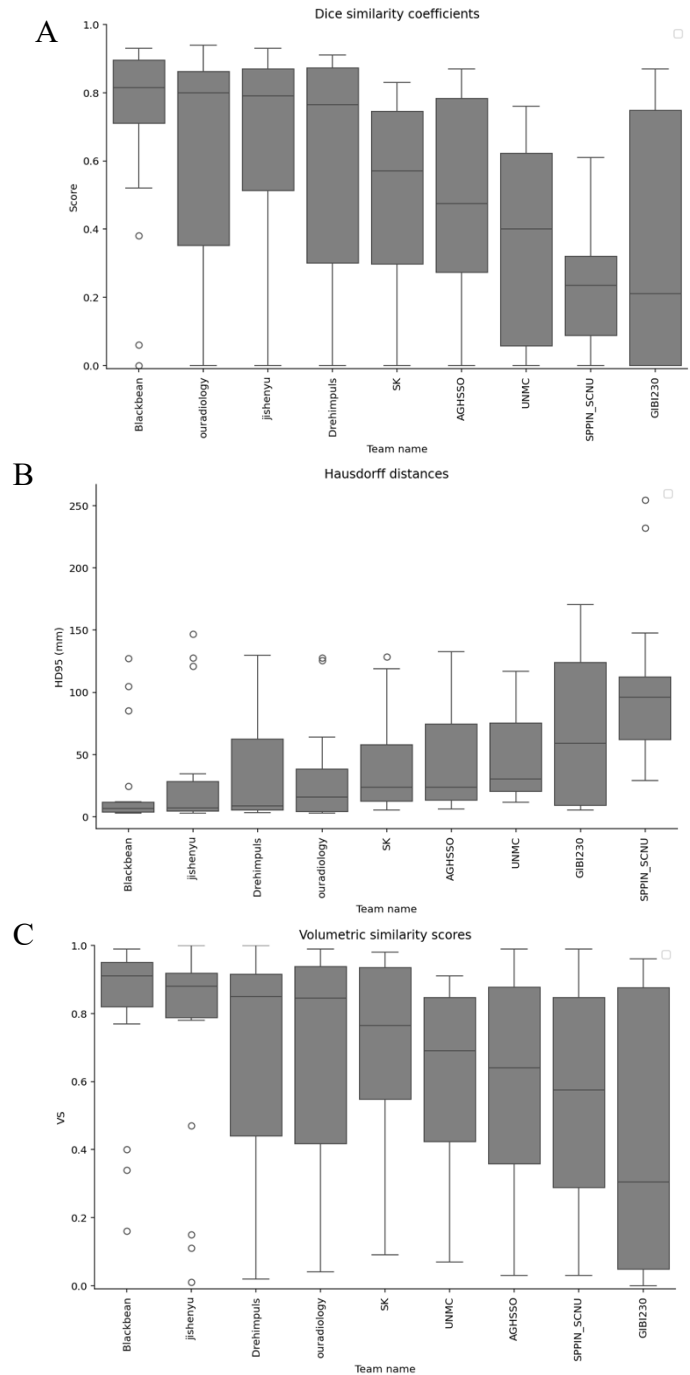

*Figure 4: The tree evaluation parameters for each team, plotted in order of the final ranking. A: the Dice similarity coefficients. B: the 95th percentile of the Hausdorff distances (HD95). Please note that HD95 values resulting from an empty segmentation are ignored while plotting. C: The volumetric similarity scores.*

### 3.4 Statistical analysis

The median Dice score for the three top scoring teams and the lowest scoring team were significantly different between the diagnostic (chemo-naïve) and post-chemotherapy scans. For the winning team Blackbean, the median score for diagnostic scans 0.81, in contrast to post-chemotherapy scans with a median Dice score of 0.47 (P = 0.01). For the HD95 only SPPIN_SCNU had a significantly lower HD95 for the diagnostic segmentations. For the VS only team jishenyu had a significant difference between the diagnostic and post-chemotherapy segmentations. An overview of the complete analysis can be seen in Table 5.

The effect of tumor size and pre-treatment is depicted in Figure 6. No clear effect of tumor size on the Dice score can be observed for each team.

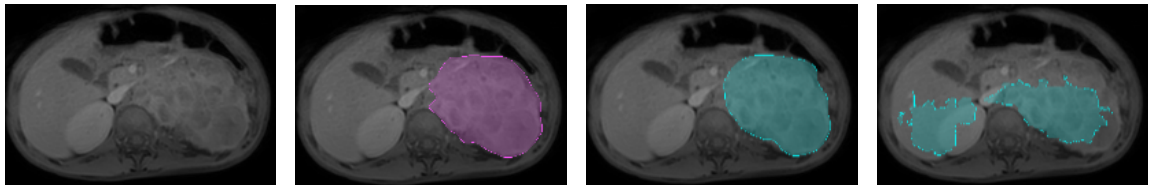

A: Highest Dice score of winning team. From left to right: $T_1$-weighted scan, ground truth, highest scoring: team Blackbean (Dice = 0.93), lowest scoring: team SPPINSCNU (Dice = 0.61). A clearly defined left-sided neuroblastoma with encasement of the vena cava and vena renalis.

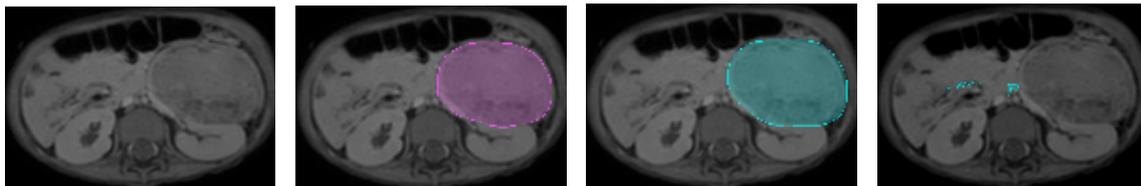

B: Highest median Dice score. From left to right: $T_1$-weighted scan, ground truth, highest scoring: team ouradiology (Dice = 0.93), lowest scoring: team SPPIN_SCNU (Dice = 0.32). This tumor is clearly defined and located in the left peritoneal space.

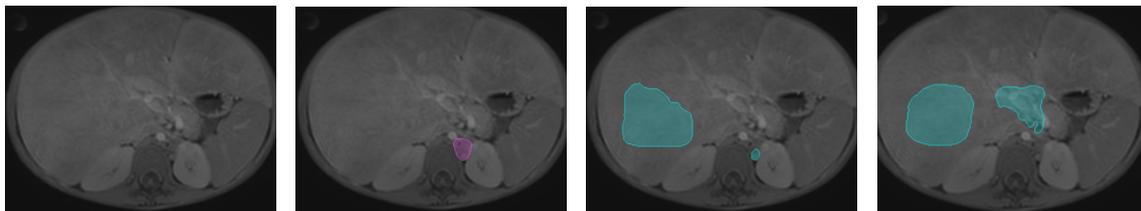

C: Lowest non-zero median Dice score. From left to right: $T_1$-weighted scan, ground truth, highest scoring: team Blackbean (Dice = 0.06), lowest scoring: team jishenyu (Dice = 0.01). This small and not well-defined primary tumor is accompanied by diffusely infiltrated liver metastases

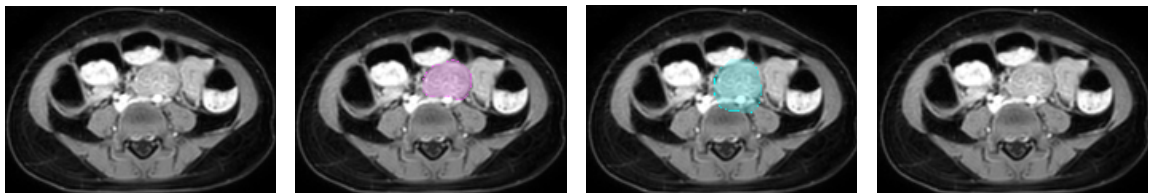

D: Lowest non-zero Dice score of Winning team. From left to right: $T_1$-weighted scan, ground truth, highest scoring: team AGHSSO (Dice = 0.65), lowest scoring: team SPPIN_ SCNU (Dice = 0.02). This is a well-defined tumor caudal to the aortic bifurcation.

*Figure 5: Overview of the segmentations of the patients belonging to the highest and lowest scored Dice scores. Image A: Test patient 9_1 (diagnostic). Image B: Test patient 5_1 (diagnostic). Image C: Test patient 3_3 (post-chemo). Image D: Test patient 1_2 (post-chemo).*

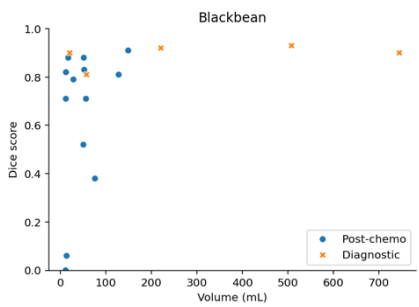
A. Team Blackbean.

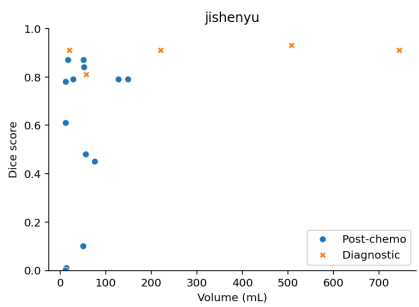
B. Team jishenyu.

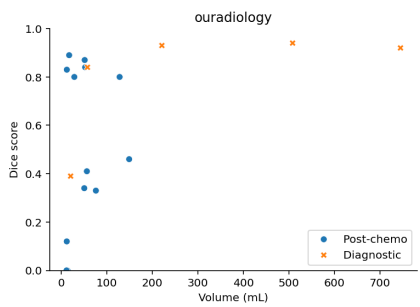
C. Team ouradiology.

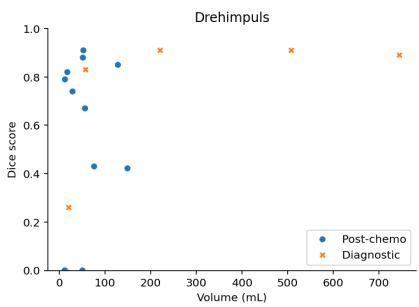
D. Team Drehimpuls.

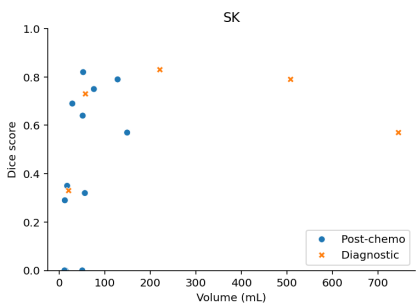
E. Team SK.

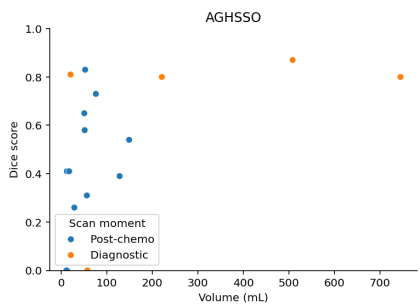
F. Team AGHSSO.

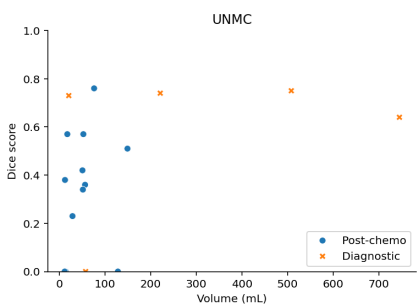
G. Team UNMC.

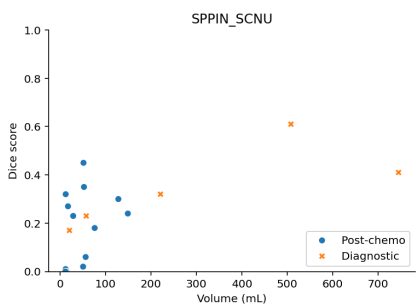
H. Team SPPIN_SCNU.

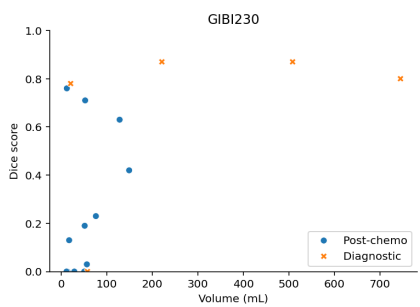
I. Team GIBI230.

Figure 6: The Dice scores plotted against the tumor volume in mL for each team. Orange crosses depict a tumor segmentation of a diagnostic MRI scan, blue points of a post-chemotherapy scan.

Table 5: The mean scores for the diagnostic and post-chemo scans for each team, using Kruskal-Wallis tests. Significant P-values (< 0.05) are depicted bold. P-values of NaN resulted from NaNs in HD scores.

| Team | Mean Dice Diagnostic | Mean Dice Post-chemo | P-Value | Mean HD Diagnostic | Mean HD post-chemo | P-Value | Mean VS Diagnostic | Mean VS Post-chemo | P-value |
|---|---|---|---|---|---|---|---|---|---|
| Blackbean | 0.89 | 0.63 | **0.01** | 6.40 | 30.17 | 0.69 | 0.94 | 0.75 | 0.05 |
| Jishenyu | 0.89 | 0.57 | **0.00** | 5.24 | 42.32 | 0.11 | 0.93 | 0.66 | **0.02** |
| Ouradiology | 0.80 | 0.51 | **0.03** | 11.51 | 39.31 | 0.20 | 0.83 | 0.61 | 0.11 |
| Drehimpuls | 0.76 | 0.50 | 0.07 | 12.55 | 48.04 | 0.40 | 0.79 | 0.62 | 0.24 |
| SK | 0.65 | 0.40 | 0.12 | 19.26 | 50.87 | 0.46 | 0.75 | 0.65 | 0.55 |
| AGHSSO | 0.66 | 0.39 | 0.07 | 31.50 | 54.11 | 0.08 | 0.79 | 0.53 | 0.08 |
| UNMC | 0.57 | 0.32 | 0.08 | 31.15 | 54.82 | 0.08 | 0.68 | 0.57 | 0.73 |
| SPPIN_SCNU | 0.35 | 0.19 | 0.15 | 53.43 | 118.93 | **0.01** | 0.51 | 0.55 | 1.00 |
| GIBI230 | 0.66 | 0.24 | **0.03** | 31.31 | 83.64 | NaN | 0.75 | 0.33 | 0.05 |

## 4. Discussion

Surgical Planning in PedIatric Neuroblastoma was the first medical imaging challenge in the field of extracranial pediatric oncology. In tot al, 9 teams were eligible for the final leaderboard. The scores of the participating teams varied widely, with the highest-ranking team achieving a median Dice score of 0.82, a median HD95 of 7.69 mm and a VS of 0.91, while the lowest-ranking team had a median Dice score of 0.21, a median HD95 of 63.31 mm and a VS of 0.31.

Although the number of teams and the variation in methods preclude general conclusions about methodology, several noteworthy observations can be made. The highest ranking team (Blackbean, Shanghai Artificial Intelligence Laboratory, Shanghai, China) used a Scalable and Transferable nnU-Net [24] as segmentation method, which is based on a nnU-Net. The unique addition of Blackbean to the nnU-Net was the extensive pre-training, on a large (> 1000 scans) dataset of CT-scans labelled with anatomical structures. This suggests that pre-training is valuable for segmentation in small, heterogenous datasets such as ours, even across modalities [32].

It is notable that most of the teams, including the top 3, used a form of nnU-Net as architecture [26]. nnU-Net has shown excellent performance on most leaderboards of recent challenges [24]. This raises the question whether investigating other deep learning methods is necessary. Drehimpuls tested several additions of nnU-Net, but they did show in their report that this did not increase performance. Other teams, such as SK, AGHSSO and UNMC, used methods other than nnU-Net, but these methods did not make top 3. As stated before, the main contribution of the highest-ranking team was the large pre-training. Together, it appears that nnU-Net is currently the best performing method and improvement for individual challenges will most likely lie in data curation, pre-training and pre-processing.

Apart from the different networks used, an important factor to consider is the input used by the teams. The majority of the teams used a single input method, with most teams using $T_1$-weighted contrast enhanced imaging as input. Several teams reported that this choice was based on the labels being created on these scans and the highest visibility of the neuroblastoma on this sequence. Only one team decided to use only $T_2$-weighted imaging as input (GIBI230). Although the method of this team showed good performance in previous publications, with Dice scores > 0.90, they ranked last in our challenge [10], [16]. As their method was trained on $T_2$-weighted imaging and no finetuning was performed on the $T_1$-weighted imaging and no registration between the $T_2$-based labels and the $T_1$-based ground truth, this indicates that their method may be of limited generalizability between sequences. Therefore, a general conclusion about the usability of $T_2$-based segmentation cannot be made based on this challenge. This also holds true when looking at the teams that used $T_2$-weighted imaging as additional input, with (n = 4) or without addition of DWI images (n = 1), because of the big spread between these teams. We selected B-values of 0 and 100 for these scans because they better preserve anatomical detail compared to higher B-values. While higher B-values provide greater sensitivity to water molecule movement and highlight areas of restricted diffusion, anatomical information is limited in these images. Nevertheless, the increased diffusion signal at higher B-values, often observed in tumors, could assist in segmenting tumors that are otherwise difficult to delineate. However, even if $T_1$-based segmentation alone is enough to provide the anatomical information needed for pre-operative planning, other sequences can be used to add additional (functional) information [33].

The highest-ranking team scored a median Dice score of 0.82, which is lower than other oncological segmentation challenges [34], [35]. When looking at the research of Veiga-Canuto about neuroblastoma segmentation, it is noticeable that their reported median Dice score of > 0.9 is significantly higher than the median Dice score of our highest-ranking team. However, Veiga-Canuto only used chemo-naïve, diagnostic MRI scans in this study, suggesting that our dataset is inherently more difficult to segment by post-chemotherapy MRI scans. Indeed, when looking at the highest Dice score of the highest-ranking team of 0.93, in a diagnostic, chemo-naïve tumor, this is in line with the reported scores in the research of Veiga-Canuto. The top three scoring teams showed a statistically significant difference in Dice score between diagnostic and post-chemotherapy scans, supporting this even more.

The inherent difficulty of our dataset is supported by the inter-observer scores for the manual segmentation, with a mean observer Dice score of 0.81. In theory, the inter-observer scores reflect the maximum scores the automated segmentation methods can score. Our inter-observer scores are low compared to the inter-observer variability of 0.97 for neuroblastic tumor segmentation as reported by Veiga-Canuto et al. [10]. This difference can partly be explained by our less experienced observers, reflected in the big spread of observer Dice scores, with a lowest median observer Dice score of 0.56. It is good to note that the segmentations in the final test set were checked and corrected by technical physicians, but to reflect the true inter-observer scores for the students only, this was not the case for the additional three segmentations used to calculate the observer Dice scores. This is reflected in the Dice score of 0 for one patient in observer 2, for which they segmented another structure instead of the tumor. Using this setup might have resulted in an underestimation of the observer Dice scores. However, for some easier tumors the median Dice score of the best scoring team is similar to the mean observer variability. In the more complex tumors, as can be seen in Figure 5, both the inter-observer variability and the team scores were lower. In conclusion, our results suggest an inherent segmentation difficulty for some scans in our dataset. Segmentation of the neuroblastoma by more experienced observers might improve the overall results of the neuroblastoma segmentation.

The spread in performance between different patients and between different tumors of the same patient is noticeable. Even the highest-ranking team had three Dice scores < 0.5. Large tumors seemed to be easy to segment, but no clear other pattern could be observed in contrast to other segmentation research [13]. Other strategies should be investigated to deal with these hard to segment, but clinically relevant cases. For example, team Drehimpuls implemented a fallback network, by doing a finetune training of the initial network by using all scans with a Dice score of < 0.5 while doing cross validation. Next, they used this so-called fallback network as a second segmentation step for patients in which < 1000 foreground pixels were detected. Despite this, their final results still included six tumors with a Dice score of < 0.5. Since small tumors typically appear on pre-operative, post-chemotherapy scan, and are challenging to segment, future research should focus on these small, yet clinically relevant tumors, by creating more extensive dataset with a focus on these tumors, investigation other MRI techniques, such as $T_1$ enhancement images and diffusion weighted imaging, and investigating other technical innovations for the segmentation [36].

Our challenge had several limitations. First, a significant number of segmentations (n = 19) failed to pass the quality check on time, which led to a smaller dataset than initially planned and potential bias for easier to segment tumors. Consequently, this also resulted in the big volume difference between the training and final test set and the preliminary test set. The large observer variability, although in part explained by the difficulty of the post-chemotherapy scans, is also a point of concern as this might have led to imperfect segmentations in the final test set and thus lower scores from the teams. Another limitation was the way of dealing with the location of the arteries and veins within segmentation of the neuroblastoma. To ensure reproducibility, our segmentation tool was as simple as possible, but this posed a challenge for the segmentation of the tumor in proximity to or encased by big vessels. Practically, only vessels close to the border of the neuroblastoma could be excluded from the tumor segmentation, but vessels fully encased by tumor were included as tumor in the ground truth. This resulted in small, but consistent errors in the ground truth tumor segmentations. However, as (manual) vessel segmentation is the next step in the workflow of creating the 3D models for neuroblastoma, this can be simply dealt with in the post-processing steps [33].

The diagnostic scans in our test set showed high segmentation performance. This is in line with the research of Veiga-Canuto, which showed Dice scores > 0.90, but only in diagnostic scans [10], [16]. Although the diagnostic scans might seem less relevant for our aim of pre-operative planning, we added these to enlarge our small database. Moreover, diagnostic MRI scans do have value in clinical planning, for example in localization of the small lesions after chemotherapy. Because this choice was reflected in our scores, with several tumors scoring low even in high performing teams, future research should focus on segmentation of these small tumors, mostly seen on pre-operative imaging. We used well established evaluation parameters for our challenge, but this might not reflect the clinical applicability completely. Further research is needed to address clinical applicability of the developed 3D models additional to image analysis evaluation parameters. Currently, the segmentation algorithms focused solely on neuroblastoma segmentation as first proof of concept. However, for a clinically applicable model, the important vessels (including but not limited to the aorta, superior mesenteric artery, inferior mesenteric artery, renal arteries, vena cava, renal veins and portal veins), and organs of interest (including kidneys, liver, spleen and pancreas) need to be included in the segmentation model. If these drawbacks are addressed, we believe that automated tumor segmentation can be a powerful tool in the surgical treatment of children with neuroblastoma.

## 5. Conclusion

The SPPIN challenge showed good results for segmentation of diagnostic tumors (Dice > 0.8), but lacked in the smaller, post-chemotherapy tumors. However, these segmentations are critical for pre-operative planning. Pretraining seems to be a good method for automated segmentation in small, heterogenous datasets present in pediatric oncology. To create clinically applicable 3D models, more reliable and extensive segmentation models are needed, with a focus on small, post-chemotherapy tumors and the inclusion of other anatomical structures.